\documentclass{INTERSPEECH2023}
\usepackage{subcaption}
\usepackage{multirow}

\interspeechcameraready

\title{End-to-End Word-Level Pronunciation Assessment with MASK Pre-training}
\name{Yukang Liang$^{1,2,^{*}\thanks{* The work was done when the first author was an intern at Microsoft Research Asia.}}$, Kaitao Song$^{3,^{+} \thanks{+ Corresponding Author.}}$, Shaoguang Mao$^3$, Huiqiang Jiang$^3$, Luna Qiu$^3$, Yuqing Yang$^3$, Dongsheng Li$^3$, Linli Xu$^{1,2}$, Lili Qiu$^3$}
\address{
  $^1$School of Computer Science and Technology, University of Science and Technology of China\\$^2$State Key Laboratory of Cognitive Intelligence $^3$Microsoft Research}
\email{liangyukang@mail.ustc.edu.cn, linlixu@ustc.edu.cn\\ \{kaitaosong, shamao, hjiang, lunaqiu, Yuqing.Yang, dongsli, liliqiu\}@microsoft.com}

\newcommand{\eg}{\emph{e.g.}}
\newcommand{\ie}{\emph{i.e.}}

\begin{document}

\maketitle
 
\begin{abstract}
Pronunciation assessment is a major challenge in the computer-aided pronunciation training system, especially at the word (phoneme)-level. To obtain word (phoneme)-level scores, current methods usually rely on aligning components to obtain acoustic features of each word (phoneme), which limits the performance of assessment to the accuracy of alignments. Therefore, to address this problem, we propose a simple yet effective method, namely \underline{M}asked pre-training for \underline{P}ronunciation \underline{A}ssessment (MPA). Specifically, by incorporating a mask-predict strategy, our MPA supports end-to-end training without leveraging any aligning components and can 
solve misalignment issues to a large extent during prediction. Furthermore, we design two evaluation strategies to enable our model to conduct assessments in both unsupervised and supervised settings. Experimental results on SpeechOcean762 dataset demonstrate that MPA could achieve better performance than previous methods, without any explicit alignment. In spite of this, MPA still has some limitations, such as requiring more inference time and reference text. They expect to be addressed in future work.
\end{abstract}




\section{Introduction}
Computer-Aided Pronunciation Training (CAPT) system~\cite{Chesta2019CAPT} is a powerful tool designed to assist humans in improving their language skills by using advanced AI technologies. Specifically, in CAPT systems, automatic Pronunciation Assessment (PA) is considered one of the two most indispensable components (the other is mispronunciation detection and diagnosis). The goal of pronunciation assessment is to analyze the characteristics of human pronunciation, and then estimate scores to reflect the quality of human pronunciation. However, building a pronunciation assessment system with high accuracy is still a challenging task in academia and industry.

Previous research~\cite{Witt2000GOP, Wenping2015mispronunciation, Wei2016Mispronunciation, Kun2017Mispronunciation, Lei2018E2EScore, Sudhakara2019GOP, Yan2020Mispronunciation, Jiatong2020GOP, Xiaoshuo2021Mispronunciation, Eesung2022SSL_scoring, Shaoguang2022UOR} on PA tasks can be broadly classified into two categories: utterance-level assessment and word-level (or phoneme-level) assessment. For utterance-level assessment, PA models are required to provide a score that accurately reflects the quality of the pronunciation. Existing works~\cite{Lei2018E2EScore, Eesung2022SSL_scoring} mainly utilized handcrafted features or deep neural networks to extract acoustic features and then aggregated these features as a final representation for utterance-level scoring. Likewise, for word-level assessment, it requires the model to provide word-level scores based on human pronunciation so that humans could better understand which parts are problematic. In consequence, word-level assessment is more difficult when compared with utterance-level assessment, since the task requires the model to capture the alignments between the audio and reference text. Therefore, existing approaches for word-level assessment generally follow a two-stage process, consisting of alignment and scoring. In these works, the most representative method is Goodness of Pronunciation (GOP)~\cite{Witt2000GOP}, which calculates the likelihood scores of the pronounced phonemes over the segmented audio by using the Gaussian mixture model-hidden Markov model and then computes the similarity with aligned phonemes. Benefiting from such design, a lot of works~\cite{Wenping2015mispronunciation, Sudhakara2019GOP, Jiatong2020GOP, Bin2021PA} attempted to improve model performance by applying more powerful acoustic models. Despite their noteworthy achievements, GOP-based methods still have several limitations. One major drawback of current GOP-based methods is a dependency on accurate audio-text alignments before generating word-level scores. Any inaccuracies in the alignment process can negatively impact the overall quality of pronunciation assessment. To avoid this problem, several studies~\cite{Leung2019Mispronunciation, Yiqing2020SEDMDD} attempted to combine both alignment and recognition together to eliminate the dependency on this audio-text alignment. For example, \cite{Leung2019Mispronunciation} utilized connectionist temporal classification (CTC) to identify phoneme sequences and obtained the corresponding acoustic features of each word (phoneme) for the final assessment by leveraging the implicit alignments of CTC. Nonetheless, these recognition-based methods still suffer from misalignment issues since these models cannot guarantee the correctness of all predicted tokens. Therefore, how to conduct word-level assessments on acoustic features with precise alignments remains a challenge. 

To address the above issues, in this paper, we propose a simple yet effective method for pronunciation assessment, which allows the model to train in an end-to-end manner, eliminating the problem of misalignment in word-level assessment. More specifically, in PA tasks, it is standard practice to provide reference texts, as human participants must read these data to gather pronunciation information. Thus, we hypothesize that the critical aspect in resolving the misalignment problem in PA tasks is how to effectively utilize the information of the reference text. Naturally, a straightforward approach is to combine audio and text features for recognition, allowing the model to learn implicit alignment during training. However, if we directly use original text data as the input for recognition, it will cause information leakage during the training and affect the subsequent prediction. Therefore, for each token prediction, we believe that the model should be able to use the whole sentence information except the predicted one. To fulfill this target, we employ an encoder-decoder framework and borrow the idea of masked language modeling~\cite{devlin2019bert} into our framework. Specifically, the encoder and the decoder are used to process audio and text data respectively, while the decoder is also bidirectional~\footnote{The decoder is equal to a bidirectional encoder plus the cross-attention module.} so that it can utilize the whole sentence information. 
During prediction, we feed audio and text data into the encoder and decoder, while only one token on the text sequence will be masked. Since other non-predicted tokens are provided, the masked token could automatically capture the dependency from the acoustic features based on the attention mechanism and thus obtain the aligned acoustic features for the final prediction. Such a strategy does not need any supervised data and enables the model to conduct assessments without the consideration of misalignment or misrecognition issues. To obtain the estimation of all tokens in the sentence, we will deploy this mask-predict operation an equal number of times as the length of the sentence. 
In addition, to enhance the generalization performance, we also incorporate pre-training into our model. More in detail, we collect public automatic speech recognition(ASR) datasets~\cite{Vassil2015Librispeech} as the pre-training corpus, and then use paired audio-text data as the inputs for the encoder and decoder. Following the experiences of BERT~\cite{devlin2019bert}, we mask a ratio of text sequence and then predict these tokens on the decoder for pre-training. Therefore, we name our model as \textbf{MPA} -- a \underline{M}asked pre-training for \underline{P}ronunciation \underline{A}ssessment. Furthermore, by providing word-level human annotations, we also design a fine-tuning strategy for our MPA, which allows our model to train in a supervised setting.

\begin{figure}[!t]
    \centering
    \begin{subfigure}[t]{0.49\textwidth}
        \centering
        \includegraphics[width=\textwidth]{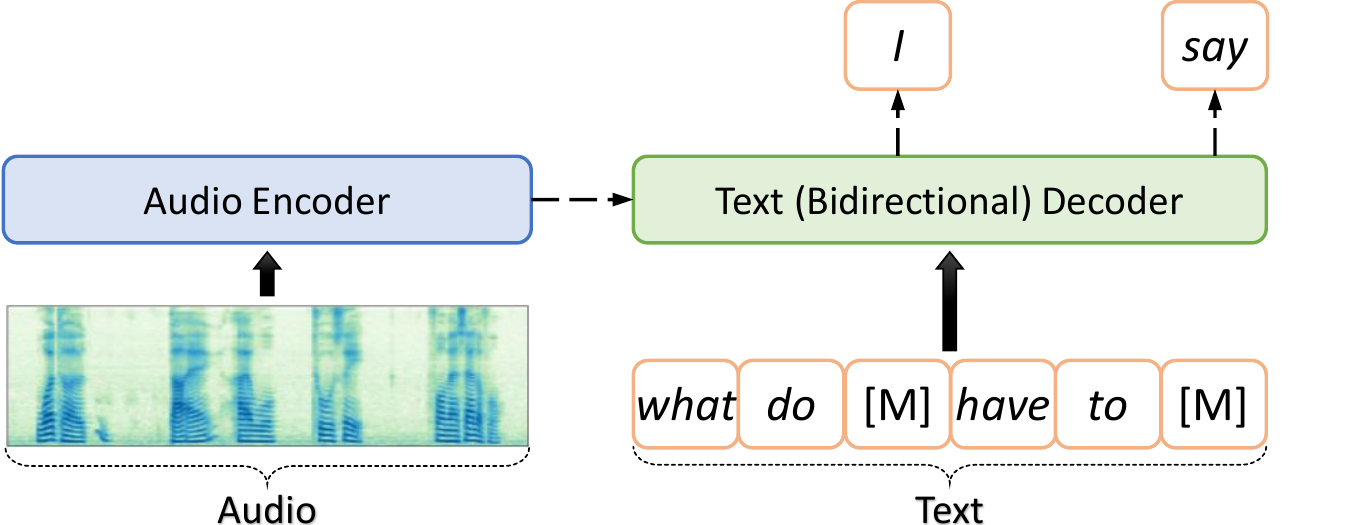}
        \caption{Pre-training}
        \label{PT}
    \end{subfigure}
    \begin{subfigure}[t]{0.49\textwidth}
        \centering
        \includegraphics[width=\textwidth]{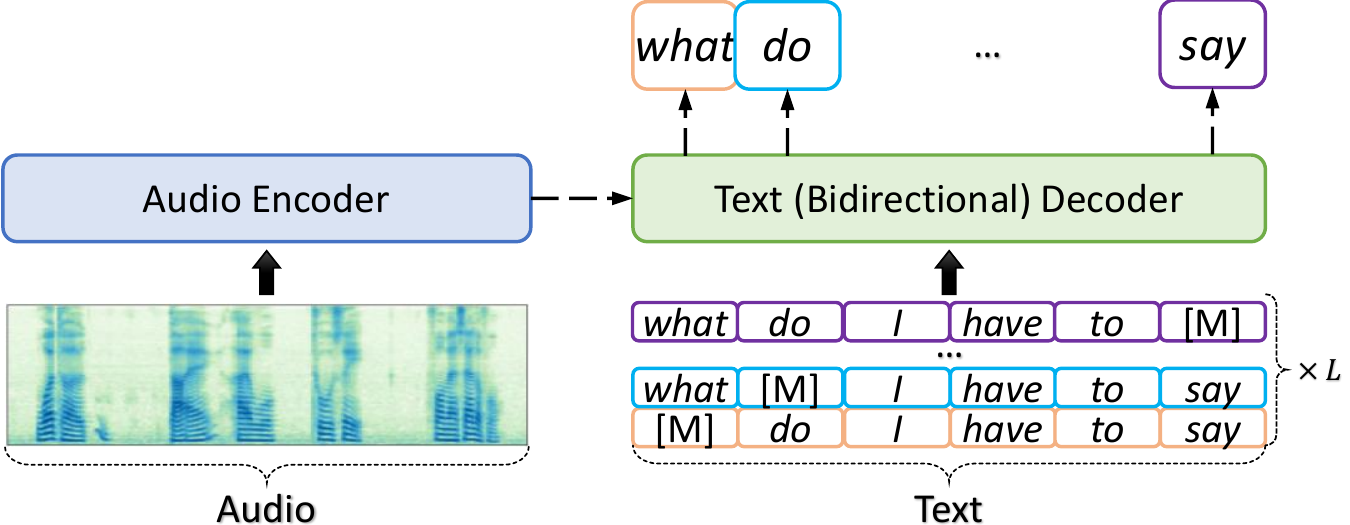}
        \caption{Unsupervised Evaluation}
        \label{unsup}
    \end{subfigure}
    \begin{subfigure}[t]{0.49\textwidth}
        \centering
        \includegraphics[width=\textwidth]{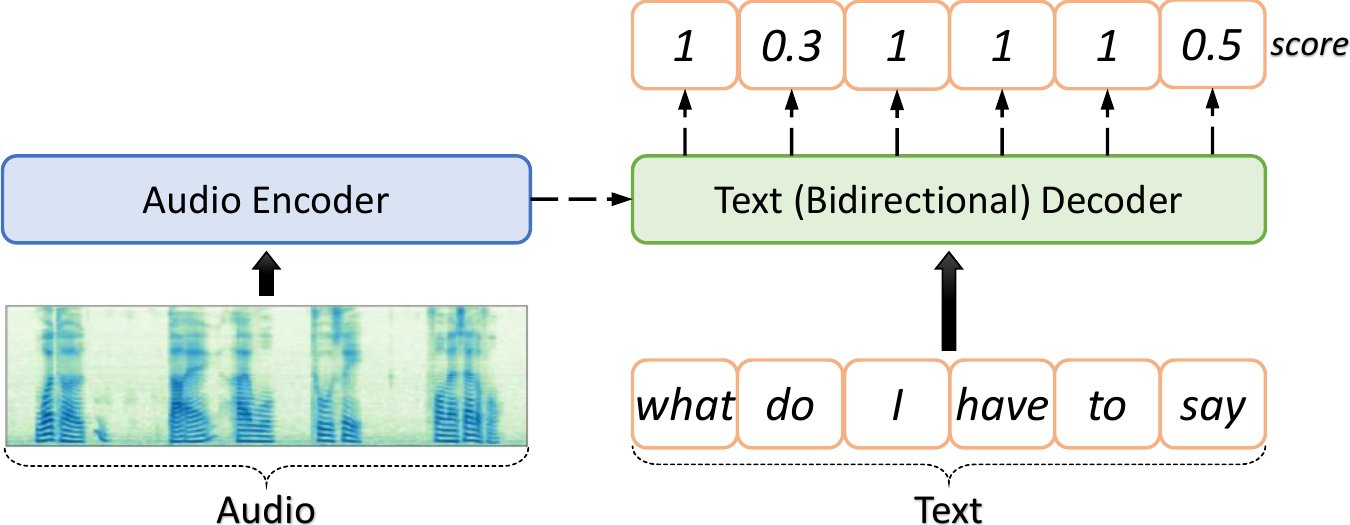}
        \caption{Fine-tuning Evaluation}
        \label{sup}
    \end{subfigure}
    \caption{The pre-training and evaluation paradigm of our proposed MPA for pronunciation assessment.}
    \label{fig:MPA}
\end{figure}

Overall, in this paper, we propose MPA, a masked pre-training method for pronunciation assessment. Our method has these advantages: 1) MPA can perfectly avoid misalignment or misrecognition problems in pronunciation assessment, by utilizing reference text and incorporating a mask-and-predict strategy for prediction; 2) MPA supports end-to-end training, as compared to two-stage paradigms; 3) We equip MPA model with two evaluation strategies, to support both unsupervised and supervised settings. Empirical results also demonstrate the effectiveness of our proposed method in pronunciation assessment.

\section{Method}
\label{sub:method}

\subsection{Pre-training}
In MPA, we adopt an encoder-decoder framework, where the encoder and the decoder are respectively used to process audio and text data. Specifically, different from the previous encoder-decoder structures~\cite{Dzmitry2015NMT, Vaswani2017Transformer}, the decoder used in MPA can be considered as a bidirectional encoder~\footnote{The standard decoder can only utilize uni-directional information, by using a left-to-right mask operation.} plus cross-attention component, similar as \cite{Marjan2019CMLM}. Based on this structure, we collect some automatic speech recognition (ASR) corpus, which includes enormous paired audio-text data, for pre-training. Denoting the paired audio and text data as $X$ and $Y$, we mask a subset of tokens $Y_{\cal{S}}$ in $Y$ and replace them as a special symbol $\rm{[MASK]}$. During the pre-training, we use the log-likelihood of predicting $Y_{\cal{S}}$ from the $X$ and $Y_{\backslash \cal{S}}$ as the objective function:
\begin{equation}
    {\cal L} (Y_{\cal{S}} | X, Y_{\backslash \cal{S}};\theta) = \sum_{i=1}^{|\cal{S}|} \log P(y_{{\cal S}_i}|X, {Y}_{\backslash \cal{S}}; \theta),
\end{equation}
where $\theta$ is the parameter of MPA. As shown in Fig.~\ref{PT}, MPA is able to utilize acoustic features and further learn bidirectional context from text features. By incorporating the pre-training paradigm, MPA is capable of achieving better generalization performance for prediction.

\subsection{Unsupervised Evaluation}
As aforementioned, how to obtain precise alignment is a key aspect of word-level pronunciation assessment. The challenges with current methods for obtaining alignment can be divided into two categories: 1) using forced alignments can lead to misalignments, and 2) using recognition-based methods to capture implicit alignments can lead to inaccuracies if the model makes incorrect predictions, such as omissions or repetitions. To obtain precise alignments, we expect the model to only estimate one token and fix other tokens for each time so that the prediction of each token can be guaranteed and the features used for prediction will automatically capture the most relevant parts from the acoustic features for the predicted one.

Therefore, based on our pre-trained MPA, we design a mask-predict strategy to evaluate pronunciations in an unsupervised setting. Specifically, for each audio $X$ and its corresponding reference text $Y$, if we want to obtain the likelihood score of the token $y_i$, we will mask this token and keep the remaining tokens and audio data. The objective function is defined as:
\begin{equation}
\label{eq2}
    {\cal L} (y_i | X, Y_{\backslash i}; \theta) = \log P(y_i|X, {Y}_{\backslash i}; \theta).
\end{equation}
By computing the above equation, we can obtain the score of a token $y_i$. To achieve the scores of all tokens in the whole sentence, we can re-compute Eqn.~\ref{eq2} with $|Y|$ times for each different token. By providing all tokens in the reference text except the token that needs to be predicted, the masked token could automatically capture which part in acoustic features should be aligned and then utilize it for prediction. Fig.~\ref{unsup} presents the details of unsupervised evaluation for MPA. 

This design of mask-predict strategy is similar to MLM-scoring~\cite{Julian2020MLMScoring}, which adopts BERT on generated text candidates for solving re-ranking tasks. While in PA tasks, we find such a design is more appropriate since PA tasks usually provide reference text for utilization. However, the proposed evaluation also retains some limitations. The main problem of the MPA unsupervised evaluation is that it needs $|Y|$ times computations, which is time-consuming. Targeting this problem, some recent works~\cite{Kaitao2022Transcormer} introduce specific solutions like sliding language modeling to address it, which we leave as future work.

\subsection{Fine-tuning Evaluation}
Based on the aforementioned design, our MPA model is able to assess the quality of pronunciation by using the log-likelihood value of the mask-predict strategy. However, some datasets could provide human annotations for subjective evaluation. Therefore, we further design a fine-tuning strategy to train our model in a supervised scenario. Specifically, we directly stack a classifier layer on the output features of the decoder in the MPA model, to measure the scores of human evaluation, as shown in Fig.~\ref{sup}. Compared with previous GOP-based or recognition-based methods which only use acoustic features, the output features of MPA are able to fully utilize both acoustic and text data, and thus provide better semantics. 

\section{Results}

\subsection{Experimental Setting}
\subsubsection{Datasets}
\paragraph*{Pre-training} we choose LibriSpeech~\cite{Vassil2015Librispeech} as the pre-training corpus, which includes nearly 960hrs audio with paired text data. For audio inputs, we first resample audios at 16,000 Hz and then extract 80-channel log mel filter-bank features (25 ms window size and 10 ms shift). For text data, we respectively process it as word level and phoneme level, to explore the performance of our model in two different settings. Following previous experience~\cite{wang2020fairseq}, we use sentencepiece~\cite{Taku2018Sentencepiece} and G2P~\footnote{\url{https://github.com/Kyubyong/g2p}} to obtain word-level and phoneme-level data. After preprocessing, the vocabulary size of word-level and phoneme-level data are 10,000 and 70, respectively. 
\paragraph*{Evaluation} For the downstream tasks, we choose SpeechOcean762~\cite{Junbo2021SpeechOcean762} as the evaluation dataset. SpeechOcean762 is a free public dataset designed for pronunciation assessment, which includes 5000 English sentences recorded by 250 non-native English speakers. This dataset provides human annotations to denote the audio quality from multiple aspects (\eg, accuracy, fluency, and prosody) at the phoneme level, word level, and sentence level. Here, in our experiments, we mainly evaluate the word-level and phoneme-level accuracy~\footnote{In SpeechOcean762, it provides human annotations to reflect word (phoneme)-level accuracy. Specifically, the phoneme-level accuracy is assigned within a range of 0 to 2, and the word-level accuracy is from a range of 0 to 10.}. The training set and test set are divided by a ratio of 5:5. We adopt the same strategy as the pre-training stage to process audio and text data.

\subsubsection{Model Configuration}
Our MPA uses an adapted Transformer~\cite{Vaswani2017Transformer} model~\footnote{\url{https://github.com/facebookresearch/fairseq/blob/main/fairseq/models/speech_to_text/s2t_transformer.py}} as the base model, which is composed of a standard Transformer encoder plus multiple convolutional subsamplers and a standard Transformer decoder~\footnote{Here, we choose \emph{s2t\_transformer} setting for model configuration.}. We train the MPA model for phoneme level and word level, which includes 47M and 57M parameters, respectively. During the pre-training, we randomly sample the number of masked tokens from a uniform distribution between one and the sequence length, and replace these tokens as a special symbol $\rm{[MASK]}$. We pre-trained our model using 4 NVIDIA Tesla V100 GPUs with a batch size of 40000 tokens per GPU and 600 epochs on the LibriSpeech dataset. All of our code is implemented based on Fairseq-S2T~\cite{wang2020fairseq}. 

\subsubsection{Metrics}
In our experiments, we choose accuracy, Mean Squared Error (MSE), and Pearson Correlation Coefficient (PCC) as the evaluation metrics.  Assuming the sequence of predicted scores is $X = \{x_1, \dots, x_n\}$ and the target sequence is ${Y} = \{{y}_1, \cdots, {y}_n\}$, MSE calculates the average of the squares of errors as:
\begin{equation}
    \rm{MSE}(X, Y) = \frac{1}{n} \sum_{i=1}^n (x_i - y_i)^2.
\end{equation}
And PCC measures the linear correlation between two sets of data, which is calculated as:
\begin{equation}
    \rm{PCC}(X, Y) = \frac{\sum_{i=1}^{n} (x_i - \bar{x}_i) \sum_{i=1}^{n} (y_i - \bar{y}_i)}{\sqrt{\sum_{i=1}^{n} (x_i - \bar{x}_i)^2} \sqrt{\sum_{i=1}^{n} (y_i - \bar{y}_i)^2}}.
\end{equation}

\begin{table}[]
    \centering
    \begin{tabular}{l|c | c c c}
    \toprule
    Model & FT & ACC $\uparrow$ & MSE $\downarrow$  & PCC $\uparrow$ \\
    \midrule 
    \multicolumn{5}{l}{\textit{Phoneme-Level}} \\
    \midrule
    GOP~\cite{Junbo2021SpeechOcean762}       & $\checkmark$ & - &  0.69 & 0.25 \\
    GOP + RF~\cite{Junbo2021SpeechOcean762}  & $\checkmark$ & - &  0.13 & 0.44 \\
    GOP + SVR~\cite{Junbo2021SpeechOcean762} & $\checkmark$ & - &  0.16 & 0.45 \\
    NNR                                      & $\checkmark$ & - &  0.35 & 0.21 \\
    UOR                                      & $\checkmark$ & - &  0.12 & 0.52 \\
    ASR                                      &              & 86.9 (\%) &  -   &   -  \\
    \midrule
    MPA$_{\rm Unsup}$ &  & 93.0 (\%) & 0.24 & 0.32 \\
    MPA$_{\rm FT}$    & $\checkmark$ & - & 0.10 & 0.89 \\
    \midrule 
    \multicolumn{5}{l}{\textit{Word-Level}} \\
    \midrule
    ASR               &  & 76.8 (\%) &  -   &   -  \\
    \midrule
    MPA$_{\rm Unsup}$ &  & 89.1 (\%) & 7.91 & 0.42 \\
    MPA$_{\rm FT}$    & $\checkmark$ & - & 2.14 & 0.94 \\
    \bottomrule
    \end{tabular}
    \caption{Performance comparison on the SpeechOcean762 dataset at the phoneme-level and word-level. The ``FT" column means whether to use labeled data for fine-tuning. The ``ACC" column means the classification accuracy of the word (phoneme) data.}
    \label{tab:phoneme_score}
\end{table}

\subsection{Performance}
We present the results of our proposed MPA at both the phoneme-level and word-level, including the unsupervised and fine-tuning settings. The baselines include 1) Goodness of Pronunciation~\cite{Witt2000GOP}; 2) GOP + Random Forest (RF); 3) GOP +  Support Vector Regressor (SVR); 4) Neural Net Regression (NNR)~\cite{Shaoguang2022UOR}; 5) Universal Ordinal Regression (UOR)~\cite{Shaoguang2022UOR}; 6) Autoregressive Automatic Speech Recognition (ASR) pre-trained on LibriSpeech. For the unsupervised setting, we use the log-likelihood value (\ie, Eqn.~\ref{eq2}) estimated by MPA to calculate ACC, MSE~\footnote{We scale the estimated log-likelihood value to a range of 0 to 2, to match the setting in SpeechOcean762.} and PCC. Since the range of word-level scores is from 0 to 10, its MSE score will be larger than phoneme-level scores. The results are shown in Table~\ref{tab:phoneme_score}. From Table~\ref{tab:phoneme_score}, we have these observations: 1) Our MPA$_{\rm Unsup}$ in the unsupervised setting can obtain a better performance in recognition than the standard ASR model, which indicates the ability of our model in prediction by utilizing full information of acoustic and text data; 2) Even without using any labeled data, our MPA$_{\rm Unsup}$ achieves a MSE score of 0.25 and a PCC score of 0.32, which can match some previous works with supervised training; 3) When fine-tuning our model with labeled data, our MPA directly obtains significant improvements  (0.32 $\rightarrow$ 0.89 and 0.42 $\rightarrow$ 0.94), which outperforms previous methods by a large margin. We guess these significant improvements are mainly brought by using additional bidirectional text features plus acoustic features. In total, all results demonstrate the effectiveness of our method in pronunciation assessment.

\subsection{Ablation Study}
\begin{table}[h]
    \begin{subtable}[h]{0.45\textwidth}
        \centering
        \begin{tabular}{l | c c c c}
        \toprule
        Range & 0-0.5 & 0.5-1.0 & 1.0-1.5 & 1.5-2.0 \\
        \midrule 
        Acc & 62.8 & 72.1 & 77.6 & 94.7 \\
        \bottomrule
       \end{tabular}
       \caption{The Accuracy of MPA at the Phoneme-level (Unsupervised).}
    \end{subtable}
    \hfill
    \begin{subtable}[h]{0.45\textwidth}
        \centering
        \begin{tabular}{l | c c c c c}
        \toprule
        Range & 0-1 & 2-3 & 4-6 & 7-9 & 10 \\
        \midrule 
        Acc & 14.3 & 19.2 & 56.8 & 77.2 & 92.8 \\
        \bottomrule
       \end{tabular}
       \caption{The Accuracy of MPA at the Word-level (Unsupervised).}
    \end{subtable}
     \caption{MPA Accuracy at different ratings.}
     \label{tab:corr}
     \vspace{-20pt}
\end{table}

\paragraph*{Correlation} To investigate if the prediction of  MPA is able to reflect the correlation with pronunciation quality, we count up the accuracy of our model at different ratings. The results are reported on speechocean762 and shown in Table~\ref{tab:corr}. We find that the accuracy of MPA is positively correlated with the word (phoneme)-level scores. These results also demonstrate that our MPA is able to reflect the quality of human pronunciation.

\begin{table}[h]
    \centering
    \begin{tabular}{c|c | c | c| c}
    \toprule
         &  AR  & Step=1 & Step=5 & Step=n \\
    \midrule
    WER  & 5.56 & 15.64  & 6.76   & 2.29 \\
    \bottomrule
    \end{tabular}
    \caption{Word Error Rate on different inference methods. The ``AR" column means using autoregressive ASR for generation. ``Step=i" means predict i steps where each step predicts tokens with a ratio of ``n/i".}
    \label{tab:infer}
    \vspace{-20pt}
\end{table}

\paragraph*{Inference Steps} Besides, we also conduct experiments to investigate the necessity of multiple inference steps. We report the word error rate (WER) of different inference strategies on librispeech dataset, and the results are reported in Table~\ref{tab:infer}. We find that by using more inference steps (\ie, each step predicts fewer tokens), the model can get better results. When using n inference steps, the accuracy of MPA could even outperform the autoregressive model. These results also indicate the effectiveness of our model to do recognition in the PA tasks, by providing reference text.

\subsection{Alignment Visualization}
\begin{figure}[!t]
    \centering
    \includegraphics[width=0.49\textwidth]{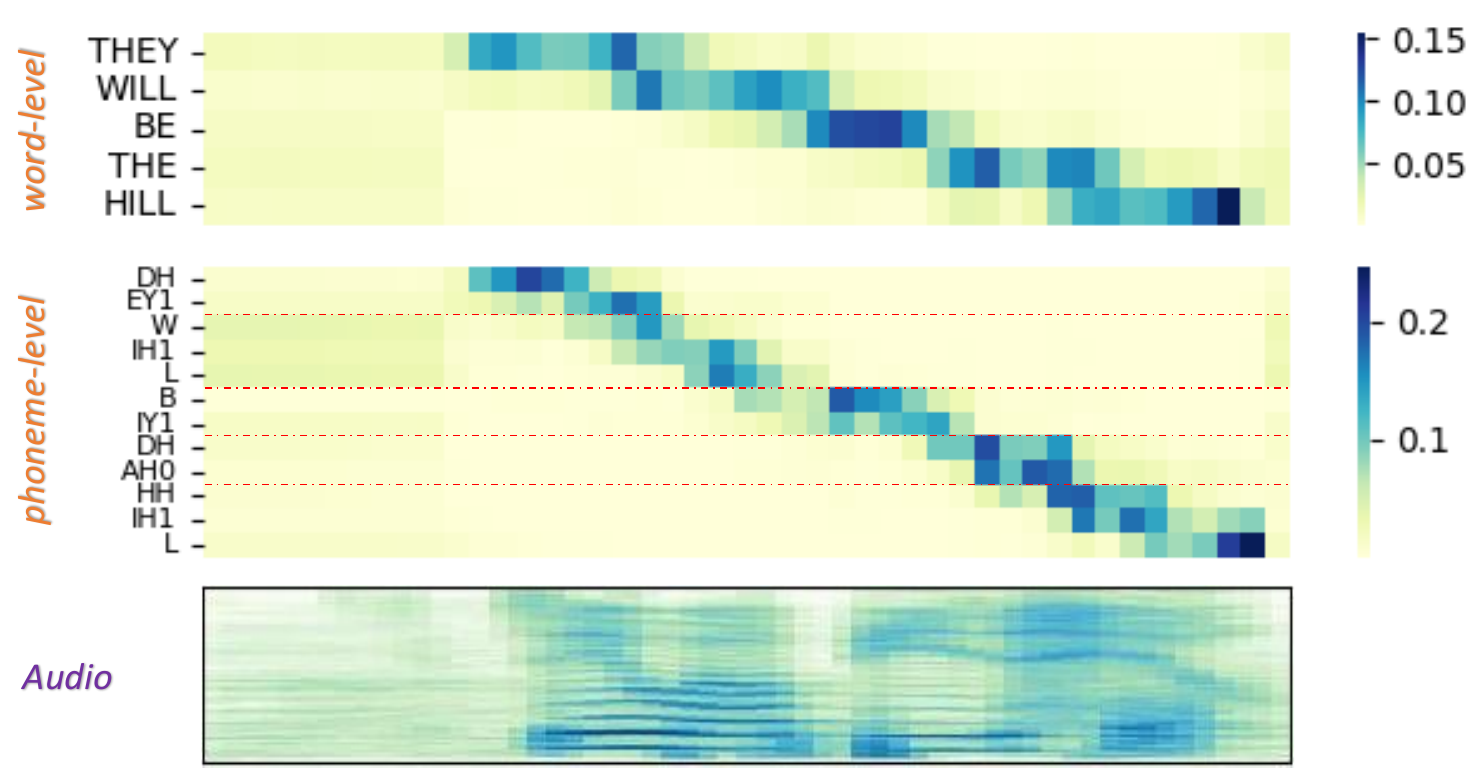}
    \caption{Attention visualization of MPA at the word level and phoneme level. The X-axis is the Mel-spectrogram of the audio, and the Y-axis is the corresponding text data (word or phoneme). The red auxiliary lines mean the alignment between word level and phoneme level.}
    \label{fig:attn}
    \vspace{-10pt}
\end{figure}

To manifest the ability of our model in alignment, we also visualize the attention weight of the last cross-attention layer. Specifically, for a sequence with $n$ tokens, we conduct the mask-predict strategy for each token and visualize the attention weight of each mask token to the audio sequence. The results are shown in Fig.~\ref{fig:attn}. We can find that the attention weight of our MPA possesses good monotonicity and each token can precisely capture the corresponding segments in the audio. Besides, the attended area at the phoneme level can also match the attended area at the word level. Since our model directly uses the reference text for alignment, we can guarantee the number of final predicted tokens is equal to the ground truth, and do not need to consider misalignment/misrecognition problems anymore.

\section{Conclusion}
In this paper, we propose MPA -- a Masked pre-training method for word (phoneme)-level Pronunciation Assessment, that aims to handle misalignment and misrecognition problems in previous works in this field, which may affect the accuracy of pronunciation assessment. Specifically, we adopt a masked pre-training method for our model on the general domain to enhance its generalization performance on the downstream tasks. During the evaluation, we utilize the content of reference text and employ a mask-predict strategy to assess the quality of each token. Furthermore, we also provide a fine-tuning setting for MPA by providing human annotations, to allow our model to train in both unsupervised and supervised settings. Benefiting from such design, our model can accurately address previous misalignment issues and utilize the features of the whole acoustic and text data to improve the performance of predictions. However, our MPA also has some limitations, such as requiring additional inference steps and dependence on reference text. In the future, we expect to address these limitations and explore better solutions to improve MPA.

\section{Acknowledgement}
This research was partially supported by the National Natural Science Foundation of China (Grant No. 62276245), and Anhui Provincial Natural Science Foundation (Grant No. 2008085J31).

\bibliographystyle{IEEEtran}
\bibliography{mybib}

\begin{thebibliography}{10}
\providecommand{\url}[1]{#1}
\csname url@samestyle\endcsname
\providecommand{\newblock}{\relax}
\providecommand{\bibinfo}[2]{#2}
\providecommand{\BIBentrySTDinterwordspacing}{\spaceskip=0pt\relax}
\providecommand{\BIBentryALTinterwordstretchfactor}{4}
\providecommand{\BIBentryALTinterwordspacing}{\spaceskip=\fontdimen2\font plus
\BIBentryALTinterwordstretchfactor\fontdimen3\font minus
  \fontdimen4\font\relax}
\providecommand{\BIBforeignlanguage}[2]{{%
\expandafter\ifx\csname l@#1\endcsname\relax
\typeout{** WARNING: IEEEtran.bst: No hyphenation pattern has been}%
\typeout{** loaded for the language `#1'. Using the pattern for}%
\typeout{** the default language instead.}%
\else
\language=\csname l@#1\endcsname
\fi
#2}}
\providecommand{\BIBdecl}{\relax}
\BIBdecl

\bibitem{Chesta2019CAPT}
C.~Agarwal and P.~Chakraborty, ``A review of tools and techniques for computer
  aided pronunciation training {(CAPT)} in english,'' \emph{Educ. Inf.
  Technol.}, vol.~24, no.~6, pp. 3731--3743, 2019.

\bibitem{Witt2000GOP}
S.~M. Witt and S.~J. Young, ``Phone-level pronunciation scoring and assessment
  for interactive language learning,'' \emph{Speech Commun.}, vol.~30, no. 2-3,
  pp. 95--108, 2000.

\bibitem{Wenping2015mispronunciation}
W.~Hu, Y.~Qian, F.~K. Soong, and Y.~Wang, ``Improved mispronunciation detection
  with deep neural network trained acoustic models and transfer learning based
  logistic regression classifiers,'' \emph{Speech Commun.}, vol.~67, pp.
  154--166, 2015.

\bibitem{Wei2016Mispronunciation}
W.~Li, S.~M. Siniscalchi, N.~F. Chen, and C.~Lee, ``Improving non-native
  mispronunciation detection and enriching diagnostic feedback with dnn-based
  speech attribute modeling,'' in \emph{ICASSP}.\hskip 1em plus 0.5em minus
  0.4em\relax {IEEE}, 2016, pp. 6135--6139.

\bibitem{Kun2017Mispronunciation}
K.~Li, X.~Qian, and H.~M. Meng, ``Mispronunciation detection and diagnosis in
  {L2} english speech using multidistribution deep neural networks,''
  \emph{{IEEE} {ACM} Trans. Audio Speech Lang. Process.}, vol.~25, no.~1, pp.
  193--207, 2017.

\bibitem{Lei2018E2EScore}
L.~Chen, J.~Tao, S.~Ghaffarzadegan, and Y.~Qian, ``End-to-end neural network
  based automated speech scoring,'' in \emph{ICASSP}.\hskip 1em plus 0.5em
  minus 0.4em\relax {IEEE}, 2018, pp. 6234--6238.

\bibitem{Sudhakara2019GOP}
S.~Sudhakara, M.~K. Ramanathi, C.~Yarra, and P.~K. Ghosh, ``{An Improved
  Goodness of Pronunciation (GoP) Measure for Pronunciation Evaluation with
  DNN-HMM System Considering HMM Transition Probabilities},'' in
  \emph{Interspeech}, 2019, pp. 954--958.

\bibitem{Yan2020Mispronunciation}
B.~Yan, M.~Wu, H.~Hung, and B.~Chen, ``An end-to-end mispronunciation detection
  system for {L2} english speech leveraging novel anti-phone modeling,'' in
  \emph{Interspeech}.\hskip 1em plus 0.5em minus 0.4em\relax {ISCA}, 2020, pp.
  3032--3036.

\bibitem{Jiatong2020GOP}
J.~Shi, N.~Huo, and Q.~Jin, ``Context-aware goodness of pronunciation for
  computer-assisted pronunciation training,'' in \emph{Interspeech}, 2020, pp.
  3057--3061.

\bibitem{Xiaoshuo2021Mispronunciation}
X.~Xu, Y.~Kang, S.~Cao, B.~Lin, and L.~Ma, ``Explore wav2vec 2.0 for
  mispronunciation detection,'' in \emph{Interspeech}.\hskip 1em plus 0.5em
  minus 0.4em\relax {ISCA}, 2021, pp. 4428--4432.

\bibitem{Eesung2022SSL_scoring}
E.~Kim, J.~Jeon, H.~Seo, and H.~Kim, ``Automatic pronunciation assessment using
  self-supervised speech representation learning,'' in \emph{Interspeech},
  2022, pp. 1411--1415.

\bibitem{Shaoguang2022UOR}
S.~Mao, F.~K. Soong, Y.~Xia, and J.~Tien, ``A universal ordinal regression for
  assessing phoneme-level pronunciation,'' in \emph{ICASSP}.\hskip 1em plus
  0.5em minus 0.4em\relax {IEEE}, 2022, pp. 6807--6811.

\bibitem{Bin2021PA}
B.~Su, S.~Mao, F.~K. Soong, Y.~Xia, J.~Tien, and Z.~Wu, ``Improving
  pronunciation assessment via ordinal regression with anchored reference
  samples,'' in \emph{ICASSP}, 2021, pp. 7748--7752.

\bibitem{Leung2019Mispronunciation}
W.~Leung, X.~Liu, and H.~Meng, ``{CNN-RNN-CTC} based end-to-end
  mispronunciation detection and diagnosis,'' in \emph{ICASSP}, 2019, pp.
  8132--8136.

\bibitem{Yiqing2020SEDMDD}
Y.~Feng, G.~Fu, Q.~Chen, and K.~Chen, ``{SED-MDD:} towards sentence dependent
  end-to-end mispronunciation detection and diagnosis,'' in \emph{ICASSP},
  2020, pp. 3492--3496.

\bibitem{devlin2019bert}
J.~Devlin, M.-W. Chang, K.~Lee, and K.~Toutanova, ``{BERT}: Pre-training of
  deep bidirectional transformers for language understanding,'' in
  \emph{NAACL}, 2019, pp. 4171--4186.

\bibitem{Vassil2015Librispeech}
V.~Panayotov, G.~Chen, D.~Povey, and S.~Khudanpur, ``Librispeech: An {ASR}
  corpus based on public domain audio books,'' in \emph{ICASSP}, 2015, pp.
  5206--5210.

\bibitem{Dzmitry2015NMT}
D.~Bahdanau, K.~Cho, and Y.~Bengio, ``Neural machine translation by jointly
  learning to align and translate,'' in \emph{ICLR}, 2015.

\bibitem{Vaswani2017Transformer}
A.~Vaswani, N.~Shazeer, N.~Parmar, J.~Uszkoreit, L.~Jones, A.~N. Gomez, L.~u.
  Kaiser, and I.~Polosukhin, ``Attention is all you need,'' in \emph{Advances
  in Neural Information Processing Systems}, vol.~30, 2017, pp. 5998--6008.

\bibitem{Marjan2019CMLM}
M.~Ghazvininejad, O.~Levy, Y.~Liu, and L.~Zettlemoyer, ``Mask-predict: Parallel
  decoding of conditional masked language models,'' in \emph{{EMNLP-IJCNLP}},
  2019, pp. 6111--6120.

\bibitem{Julian2020MLMScoring}
J.~Salazar, D.~Liang, T.~Q. Nguyen, and K.~Kirchhoff, ``Masked language model
  scoring,'' in \emph{ACL}, 2020, pp. 2699--2712.

\bibitem{Kaitao2022Transcormer}
K.~Song, Y.~Leng, X.~Tan, Y.~Zou, T.~Qin, and D.~Li, ``Transcormer: Transformer
  for sentence scoring with sliding language modeling,'' in \emph{Advances in
  Neural Information Processing Systems ({NeurIPS})}, 2022.

\bibitem{wang2020fairseq}
C.~Wang, Y.~Tang, X.~Ma, A.~Wu, D.~Okhonko, and J.~Pino, ``Fairseq {S}2{T}:
  Fast speech-to-text modeling with fairseq,'' in \emph{Proceedings of the 1st
  Conference of the Asia-Pacific Chapter of the Association for Computational
  Linguistics and the 10th International Joint Conference on Natural Language
  Processing: System Demonstrations}, dec 2020, pp. 33--39.

\bibitem{Taku2018Sentencepiece}
T.~Kudo and J.~Richardson, ``Sentencepiece: {A} simple and language independent
  subword tokenizer and detokenizer for neural text processing,'' in
  \emph{{EMNLP} 2018: System Demonstrations}, 2018, pp. 66--71.

\bibitem{Junbo2021SpeechOcean762}
J.~Zhang, Z.~Zhang, Y.~Wang, Z.~Yan, Q.~Song, Y.~Huang, K.~Li, D.~Povey, and
  Y.~Wang, ``speechocean762: An open-source non-native english speech corpus
  for pronunciation assessment,'' in \emph{Interspeech}, 2021, pp. 3710--3714.

\end{thebibliography}

\end{document}